\documentclass[11pt,a4paper]{article}
\usepackage[hyperref]{emnlp2020}
\usepackage{times,latexsym}
\usepackage{url}
\usepackage[T1]{fontenc}

\usepackage{amsfonts}
\usepackage{comment}
\usepackage{tikz}
\usepackage{pgfplots}
\usetikzlibrary{shapes, positioning, decorations.pathreplacing, shapes.multipart}
\usepackage{amsmath}
\usepackage{todonotes}
\usepackage{algorithm}
\usepackage{multirow}
\usepackage{subfig}
\usepackage[noend]{algpseudocode}
\usepackage{booktabs}
\usepackage{mymathdef}
\usepackage[utf8]{inputenc}
\usepackage{CJKutf8}
\usepackage{threeparttable}
\usepackage{tikz}
\usepackage{tikz-dependency}
\usepackage{tabularx}
\usepackage{xparse}
\usepackage{xspace}
\usepackage{relsize}
\usepackage{graphicx}
\usepackage{multirow}
\usepackage[normalem]{ulem}
\usepackage{amsmath}
\usepackage{colortbl}

\newcommand{\tradc}[1]{\begin{CJK*}{UTF8}{bsmi}\textcolor[rgb]{1.00,0.00,0.00}{#1}\end{CJK*}}

\usepackage{microtype}

\aclfinalcopy % Uncomment this line for the final submission
\newenvironment{itemize*}%
  {\begin{itemize}%
    \setlength{\itemsep}{1pt}%
    \setlength{\parskip}{1pt}}%
  {\end{itemize}}
\newenvironment{enumerate*}%
  {\begin{enumerate}%
    \setlength{\itemsep}{1pt}%
    \setlength{\parskip}{1pt}}%
  {\end{enumerate}}

\newenvironment{enumerate**}%
  {\begin{enumerate}%
    \setlength{\itemsep}{0pt}%
    \setlength{\parskip}{0pt}}%
  {\end{enumerate}}

\title{A Concise Model for Multi-Criteria Chinese Word Segmentation with Transformer Encoder}

\author{
Xipeng Qiu\thanks{\ \  Corresponding author.}, Hengzhi Pei, Hang Yan, Xuanjing Huang \\
Shanghai Key Laboratory of Intelligent Information Processing, Fudan University \\
School of Computer Science, Fudan University \\
\{xpqiu, hzpei16, hyan19, xjhuang\}@fudan.edu.cn
}

\date{}
\begin{document}

\begin{CJK*}{UTF8}{gbsn}
\maketitle

\begin{abstract}
Multi-criteria Chinese word segmentation (MCCWS) aims to exploit the relations among the multiple heterogeneous segmentation criteria and further improve the performance of each single criterion. Previous work usually regards MCCWS as different tasks, which are learned together under the multi-task learning framework. In this paper, we propose a concise but effective unified model for MCCWS, which is fully-shared for all the criteria. By leveraging the powerful ability of the Transformer encoder, the proposed unified model can segment Chinese text according to a unique criterion-token indicating the output criterion. Besides, the proposed unified model can segment both simplified and traditional Chinese and has an excellent transfer capability. Experiments on eight datasets with different criteria show that our model outperforms our single-criterion baseline model and other multi-criteria models.
Source codes of this paper are available on Github\footnote{\url{https://github.com/acphile/MCCWS}}.
\end{abstract}

%\begin{keyword}
%Chinese Word Segmentation, Multi-Criteria, Self-Attention, Transformer Encoder, Natural Language Processing, Deep Learning
%\end{keyword}
\section{Introduction}

%Unlike English, there is no apparent delimiter between words in Chinese sentences. Since words are usually regarded as the minimum semantic units,
Chinese word segmentation (CWS) is a preliminary step to process Chinese text.
The mainstream CWS methods regard CWS as a character-based sequence labeling problem, in which each character is assigned a label to indicate its boundary information.
Recently, various neural models have been explored to reduce efforts of the feature engineering ~\citep{chen2015gated,chen2015long,yan2020jcst-semicrf,wang2017convolutional,kurita2017neural,ma2018state}.
%Although these methods have made significant progress, they considerably rely on large-scale and high-quality annotated corpus.

%However, there are several inconsistent criteria to segment Chinese sentences according to the different linguistic perspectives or different segmentation granularities. The several existing CWS corpora adopt different segmentation criteria, resulting that their segmentations for one sentence are usually inconsistent.

Recently, \citet{chen2017adversarial} proposed multi-criteria Chinese word segmentation (MCCWS) to effectively utilize the heterogeneous resources with different segmentation criteria. Specifically, they regard each segmentation criterion as a single task under the framework of multi-task learning, where a shared layer is used to extract the criteria-invariant features, and a private layer is used to extract the criteria-specific features.

However, it is unnecessary to use a specific private layer for each criterion. These different criteria often have partial overlaps. For the example in Table \ref{tab:example}, the segmentation of ``林丹 (Lin Dan)'' is the same in CTB and MSRA criteria, and the segmentation of ``总$|$冠军 (the championship)'' is the same in PKU and MSRA criteria. All these three criteria have the same segmentation for the word ``赢得 (won)''.
Although these criteria are inconsistent, they share some partial segmentation. Therefore, it is interesting to use a unified model for all the criteria. At the inference phase, a criterion-token is taken as input to indicate the predict segmentation criterion. Following this idea, \citet{gong2018switch} used multiple LSTMs and a criterion switcher at every position to automatically switch the routing among these LSTMs. \citet{he2019effective} used a shared BiLSTM to deal with all the criteria by adding two artificial tokens at the beginning and end of an input sentence to specify the target criterion. However, due to the long-range dependency problem, BiLSTM is hard to carry the criterion information to each character in a long sentence.

\begin{table}[t]\small
\centering
\begin{tabular}{|c|*{5}{c|}}
\hline
Corpora&Lin&Dan&won&\multicolumn{2}{c|}{the championship}\\
\hline
CTB&\multicolumn{2}{c|}{\cellcolor[rgb]{0.2,0.9,0.7}林丹}&赢得&\multicolumn{2}{c|}{总冠军}\\
\hline
PKU&林&丹&赢得&\cellcolor[rgb]{1,0.8,0.8}总&\cellcolor[rgb]{1,0.8,0.8}冠军\\\hline
MSRA&\multicolumn{2}{c|}{\cellcolor[rgb]{0.2,0.9,0.7}林丹}&赢得&\cellcolor[rgb]{1,0.8,0.8}总&\cellcolor[rgb]{1,0.8,0.8}冠军\\
\hline
\end{tabular}
\caption{Illustration of different segmentation criteria.}\label{tab:example}
\end{table}

\begin{figure}
  \centering
  \subfloat[CTB]{
  \includegraphics[width=0.35\textwidth]{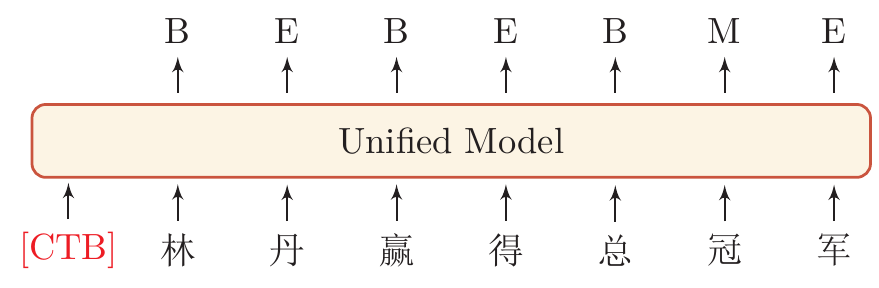} \label{fig:exam-ctb}
  }\\
  \vspace{-1em}
  \subfloat[PKU]{
  \includegraphics[width=0.38\textwidth]{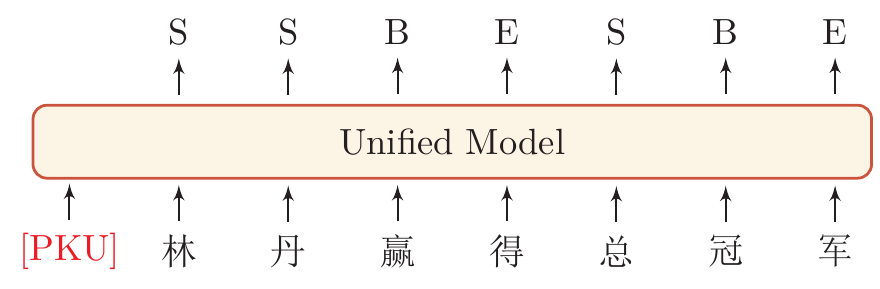} \label{fig:exam-pku}
  }
  \caption{Unified model for MCCWS. ``[$\cdot$]'' is a special token indicating the output criterion. The label $\{B, M, E, S\}$ of each character indicates it is the begin, middle, end of a word, or a word with single character.
  }\label{fig:exam}
\end{figure}

In this work, we propose a concise unified model for MCCWS task by integrating shared knowledge from multiple segmentation criteria. Inspired by the success of the Transformer~\cite{vaswani2017attention}, we design a fully shared architecture for MCCWS, where a shared Transformer encoder is used to extract the criteria-aware contextual features, and a shared decoder is used to predict the criteria-specific labels. An artificial token is added at the beginning of the input sentence to determine the output criterion. The similar idea is also used in the field of machine translation, \citet{johnson2017google} used a single model to translate between multiple languages.
Figure \ref{fig:exam} illustrates our model. There are two reasons to use the Transformer encoder for MCCWS. The primary reason is its neatness and ingenious simplicity to model the criterion-aware context representation for each character. Since the  Transformer encoder uses self-attention mechanism to capture the interaction each two tokens in a sentence, each character can immediately perceive the information of the criterion-token as well as the context information. The secondary reason is that the Transformer encoder has potential advantages in capturing the long-range context information and having a better parallel efficiency than the popular LSTM-based encoders.
Finally, we exploit the eight segmentation criteria on the five simplified Chinese and three traditional Chinese corpora. Experiments show that the proposed model is effective in improving the performance of MCCWS.
%We also observe that traditional Chinese could benefit from incorporating knowledge from simplified Chinese.

The contributions of this paper could be summarized as follows.
\begin{itemize*}
%  \item Multi-criteria learning is introduced for CWS, which aims to make full use of the existing heterogeneous corpora. %Although the segmentation criteria of these corpora are different, they share lots of common knowledge and could help each other.
  \item We proposed a concise unified model for MCCWS based on Transformer encoder, which adopts a single fully-shared model to segment sentences with a given target criterion. It is attractive in practice to use a single model to produce multiple outputs with different criteria.
  \item By a thorough investigation, we show the feasibility of using a unified CWS model to segment both simplified and traditional Chinese (see Sec. \ref{sec:bothlan}). We think it is a promising direction for CWS to exploit the collective knowledge of these two kinds of Chinese.
  \item The learned criterion embeddings reflect the relations between different criteria, which make our model have better transfer capability to a new criterion (see Sec.  \ref{sec:transfer}) just by finding a new criterion embedding in the latent semantic space.
  \item It is a first attempt to train the Transformer encoder from scratch for CWS task. Although we mainly address its conciseness and suitability for MCCWS in this paper and do not intend to optimize a specific Transformer encoder for the single-criterion CWS (SCCWS), we prove that the Transformer encoder is also valid for SCCWS. The potential advantages of the Transformer encoder are that it can effectively extract the long-range interactions among characters and has a better parallel ability than LSTM-based encoders.
\end{itemize*}

\begin{figure*}
  \centering
  \vspace{-2em}
  \subfloat[SCCWS]{
  \hspace{2em}
  \includegraphics[height=9em]{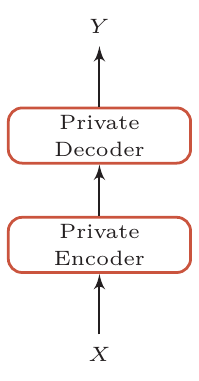} \label{fig:Model-I}
  \hspace{2em}
  }
  \subfloat[MTL-based MCCWS]{
  \includegraphics[height=9em]{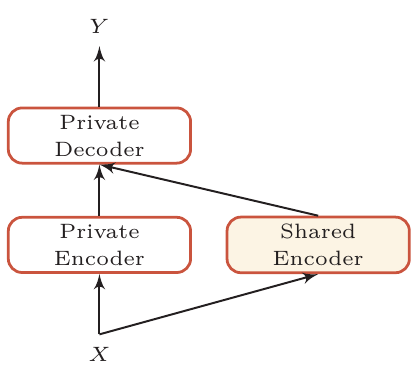} \label{fig:Model-II}
  }
  \subfloat[Unified MCCWS]{
  \hspace{4em} \includegraphics[height=9em]{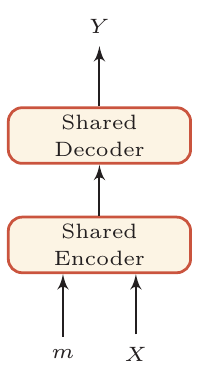} \label{fig:Model-III}
  \hspace{4em}
  }
  \caption{Architectures of SCCWS and MCCWS. The shaded components are shared for different criteria.%, and those in green are private.
  }\label{fig:cws_models}
\end{figure*}

\section{Background}

In this section, we first briefly describe the background knowledge of our work.

\subsection{Neural Architecture for CWS}

Usually, CWS task could be viewed as a character-based sequence labeling problem. Specifically, each character in a sentence $X = \{x_1, \dots, x_T\}$ is labelled as one of $y \in \cL=\{B, M, E, S\}$, indicating the begin, middle, end of a word, or a word with single character. The aim of CWS task is to figure out the ground truth of labels $Y^* = \{y_1^*, \dots, y_T^*\}$:
\begin{equation}
Y^* = \argmax_{Y \in \cL^T} p (Y | X). \label{eq:argmax}
\end{equation}

Recently, various neural models have been widely used in CWS and can effectively reduce the efforts of feature engineering.
The modern architecture of neural CWS usually consists of three components:
%(1) A character embedding layer to map the Chinese characters into dense vectors.  (2) An encoding layer to extract the contextual features, which is usually implemented by several classical neural networks. Usually, the recurrent neural networks (RNNs) or convolutional neural networks (CNNs) are adopted as the encoding layer. (3) A decoding layer to predict the labels, which is a conditional random field (CRF) \cite{lafferty2001conditional} layer or a multi-layer perceptron (MLP).
%,ye2019improving

{\bf Embedding Layer:}
In neural models, the first step is to map discrete language symbols into distributed embedding space. Formally, each character $x_t$ is mapped as $\be_{x_t} \in \mathbb{R}^{d_e}$, where $d_e$ is a hyper-parameter indicating the size of character embedding.%,

{\bf Encoding Layer:}
The encoding layer is to extract the contextual features for each character.

For example, a prevalent choice for the encoding layer is the bi-directional LSTM (BiLSTM) \cite{hochreiter1997long}, which could incorporate information from both sides of sequence.
\begin{equation}\small
\bh_t = \BiLSTM(\be_{x_t}, \overrightarrow{\bh}_{t-1}, \overleftarrow{\bh}_{t+1}, \theta_e),\label{eq:LSTM}
\end{equation}
where $\overrightarrow{\bh}_t$ and $\overleftarrow{\bh}_t$ are the hidden states at step $t$ of the forward and backward LSTMs respectively, $\theta_e$ denotes all the parameters in the BiLSTM layer.

Besides BiLSTM, CNN is also alternatively used to extract features.

{\bf Decoding Layer:}
The extracted features are then sent to conditional random fields (CRF) \cite{lafferty2001conditional} layer or multi-layer perceptron (MLP) for tag inference.

When using CRF as decoding layer, $p (Y | X)$ in Eq (\ref{eq:argmax}) could be formalized as:
\begin{equation}\small
p (Y | X) = \frac{\Psi (Y | X)}{\sum_{Y^\prime \in \cL^n} \Psi (Y^\prime | X)},
\end{equation}
where $\Psi (Y | X)$ is the potential function. In first order linear chain CRF, we have:
\begin{gather}\small
\Psi (Y | X) = \prod_{t = 2}^n \psi (X, t, y_{t-1}, y_t),\\
\psi (\bx, t, y^\prime, y) = \exp(\delta(X, t)_{y} + \bb_{y^\prime y}),
\end{gather}
where $\bb_{y^\prime y} \in \bR$ is trainable parameters respective to label pair $(y^\prime, y)$, score function $\delta(X, t) \in \mathbb{R}^{|\cL|}$ calculates scores of each label for tagging the $t$-th character:
\begin{equation}\small\small
\delta(X, t) = \bW_{\delta}^\top \mathbf{h}_t + \bb_{\delta}, \label{eq:score}
\end{equation}
where $\mathbf{h}_t$ is the hidden state of encoder at step $t$, $\bW_{\delta} \in \mathbb{R}^{d_h \times |\mathcal{L}|}$ and $\bb_{\delta} \in \mathbb{R}^{|\mathcal{L}|}$ are trainable parameters.

When using MLP as decoding layer, $p (Y | X)$ in Eq (\ref{eq:argmax}) is directly predicted by a MLP with softmax function as output layer.
\begin{equation}\small
p (y_t | X) = \mathrm{MLP}(\mathbf{h}_t,\theta_d),\qquad \forall t\in[1,T]
\end{equation}
where $\theta_d$ denotes all the parameters in MLP layer.

Most current state-of-the-art CWS models \cite{chen2015gated,xu2016dependency,liu2016exploring,yang2018subword,yan2020jcst-semicrf} mainly focus on single-criterion CWS (SCCWS).  Figure \ref{fig:Model-I} shows the architecture of SCCWS.

%chen2016neural

\subsection{MCCWS with Multi-Task Learning}
To improve the performance of CWS by exploiting multiple heterogeneous criteria corpora, \citet{chen2017adversarial} utilize the multi-task learning framework to model the shared information among these different criteria.

Formally, assuming that there are $M$ corpora with heterogeneous segmentation criteria, we refer $\mathcal{D}_m$ as corpus $m$ with $N_m$ samples:
\begin{equation}
\mathcal{D}_m = \{(X_n^{(m)},Y_n^{(m)})\}_{n=1}^{N_m},
\end{equation}
where $X_n^{(m)}$ and $Y_n^{(m)}$ denote the $n$-th sentence and the corresponding label in corpus $m$ respectively.

The encoding layer introduces a shared encoder to mine the common knowledge across multiple corpora, together with the original private encoder. The architecture of MTL-based MCCWS is shown in Figure \ref{fig:Model-II}.

Concretely, for corpus $m$, a shared encoder and a private encoder are first used to extract the criterion-agnostic and criterion-specific features.
\begin{align}
\bH^{(s)} =& \mathbf{enc}_s(\be_{X};\theta^{(s)}_e),\\
\bH^{(m)} =& \mathbf{enc}_m(\be_{X};\theta^{(m)}_e), \quad \forall m\in [1,M]
\end{align}
where $\be_{X}=\{\be_{x_1},\cdots,\be_{x_T}\}$ denotes the embeddings of the input characters $x_1,\cdots,x_T$, $\mathbf{enc}_s(\cdot)$ represents the shared encoder and $\mathbf{enc}_m(\cdot)$ represents the private encoder for corpus $m$; $\theta^{(s)}_e$ and $\theta^{(m)}_e$ are the shared and private parameters respectively. The shared and private encoders are usually implemented by the RNN or CNN network.

Then a private decoder is used to predict criterion-specific labels.
For the $m$-th corpus, the probability of output labels is
\begin{align}
p_m(Y|X) &= \mathbf{dec}_m([\bH^{(s)};\bH^{(m)}];\theta^{(m)}_d),
\end{align}
where $\mathbf{dec}_m(\cdot)$ is a private CRF or MLP decoder for corpus $m (m\in [1,M])$, taking the shared and private features as inputs; $\theta^{(m)}_d$ is the parameters of the $m$-th private decoder.

\paragraph{Objective}
The objective is to maximize the log likelihood of true labels on all the corpora:

\vspace{-1em}
{\footnotesize\begin{align}
\mathcal{J}_{seg}(\Theta^{m},\Theta^{s}) = \sum_{m=1}^{M} \sum_{n=1}^{N_m}\log p_m(Y^{(m)}_n|X^{(m)}_n;\Theta^{m},\Theta^{s}),\label{eq:objective}
\end{align}}%
where $\Theta^{m}$ = $\{\theta^{(m)}_e,\theta^{(m)}_d\}$ and $\Theta^{s}$ = $\{\bE,\theta^{(s)}_e\}$ denote all the private and shared parameters respectively; $\bE$ is the embedding matrix.

%Corpora are trained iteratively with one batch after another.

\section{Proposed Unified Model}

\begin{figure}[t]
    \centering
    \includegraphics[width=0.45\textwidth]{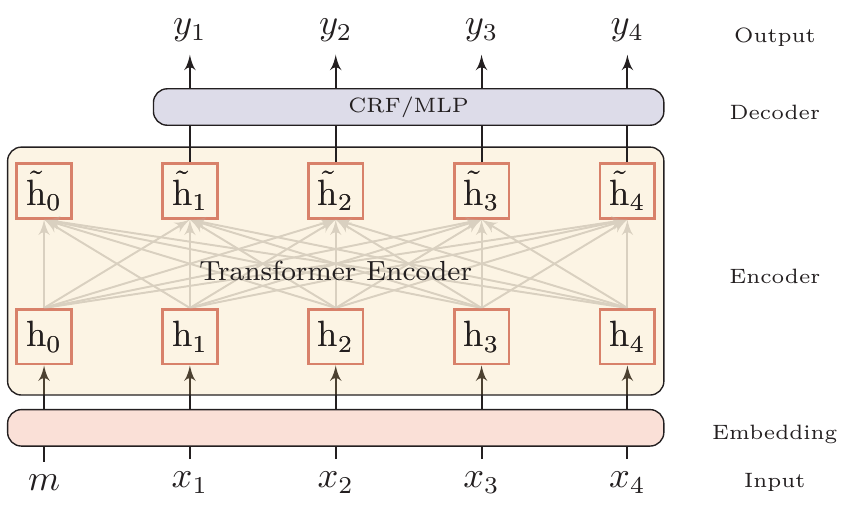}
    \caption{Proposed Model for MCCWS.}\label{fig:arch}
  \end{figure}

In this work, we propose a more concise architecture for MCCWS, which adopts the Transformer encoder~\cite{vaswani2017attention} to extract the contextual features for each input character. In our proposed architecture, both the encoder and decoder are shared by all the criteria. The only difference for each criterion is that a unique token is taken as input to specify the target criterion, which makes the shared encoder to capture the criterion-aware representation. Figure \ref{fig:cws_models} illustrates the difference between our proposed model and the previous models.
A more detailed architecture for MCCWS is shown in Figure \ref{fig:arch}.

\subsection{Embedding Layer}

Given a sentence $X = \{x_1, \dots, x_T\}$, we first map it into a vector sequence where each token is a $d_{model}$ dimensional vector. Besides the standard character embedding, we introduce three extra embeddings: criterion embedding, bigram embedding, and position embedding.

1) \textbf{Criterion Embedding:} Firstly, we add a unique criterion-token at the beginning of $X$ to indicate the output criterion.  For the $m$-th criterion, the criterion-token is $[m]$. We use $\mathbf{e}_{[m]}$ to denote its embedding. Thus, the model can learn the relations between different criteria in the latent embedding space.

2) \textbf{Bigram Embedding:} Based on~\cite{chen2015long,shao2017character,zhang2018simple}, the character-level bigram features can significantly benefit the task of CWS. Following their settings, we also introduce the bigram embedding to augment the character-level unigram embedding.
The representation of character $x_t$ is
\begin{align}
\be'_{x_{t}} = FC(\be_{x_{t}}\oplus \be_{x_{t-1}x_{t}}\oplus \be_{x_{t}x_{t+1}}),
\end{align}
where $\be$ denotes the $d$-dimensional embedding vector for the unigram and bigram, $\oplus$ is the concatenation operator, and FC is a fully connected layer to map the concatenated character embedding with the dimension $3d$ into the embedding $\be'_{x_{t}}\in \mathbb{R}^{d_{model}}$.

3) \textbf{Position Embedding:} To capture the order information of a sequence, a position embedding $PE$ is used for each position. The position embedding can be learnable parameters or predefined. In this work, we use the predefined position embedding following \cite{vaswani2017attention}. For the $t$-th character in a sentence, its position embedding is defined by
\begin{align}
PE_{t,2i} =& \sin(t/10000^{2i/d_{model}}),\\
PE_{t,2i+1} =& \cos(t/10000^{2i/d_{model}}),
\end{align}
where $i$ denotes the dimensional index of position embedding.

Finally, the embedding matrix of the sequence $X = \{x_1, \cdots, x_T\}$ with criterion $m$ is formulated as

\vspace{-1em}
{\small
\begin{align}
\bH&=[\be_{[m]}+PE_0;\be'_{x_1}+PE_1;\cdots;\be'_{x_T}+PE_T],
\end{align}}%
where $\bH\in \mathbb{R}^{(T+1)\times d_{model}}$,  $(T+1)$ and $d_{model}$ represent the length and the dimension of the input vector sequence.

\subsection{Encoding Layer}\label{sec:enc}

In sequence modeling, RNN and CNN often suffer from the long-term dependency problem and cannot effectively extract the non-local interactions in a sentence. Recently,
the fully-connected self-attention architecture, such as Transformer \cite{vaswani2017attention}, achieves great success in many NLP tasks.

In this work, we adopt the Transformer encoder as our encoding layer, in which several multi-head self-attention layers are used to extract the contextual feature for each character.

%In each multi-head self-attention layer, we use the scaled dot-product attention to model the intra-interactions of a sequence.
Given a sequence of vectors $H \in \mathbb{R}^{(T+1)\times d_{model}}$, a single-head self-attention projects $H$ into three different matrices: the query matrix $Q\in \mathbb{R}^{(T+1)\times d_{k}}$, the key matrix $K\in \mathbb{R}^{(T+1)\times d_{k}}$ and the value matrix $V\in \mathbb{R}^{(T+1)\times d_{v}}$, and uses scaled dot-product attention to get the output representation.
\begin{align}
    &Q,K,V = H W^Q, H W^K,H W^V \label{eq:qkv}\\
    &\mathrm{Attn}(Q,K,V) = \mathrm{softmax}(\frac{Q K^T}{\sqrt{d_k}}) V,
\end{align}
where the matrices $ W^Q\in \mathbb{R}^{ d_{model}\times d_k},W^K\in \mathbb{R}^{ d_{model}\times d_k},W^V \in \mathbb{R}^{ d_{model}\times d_v}$ are learnable parameters and $\mathrm{softmax}(\cdot)$ is performed row-wise.

%The multi-head self-attention is an extension of the single-head self-attention to model the multiple interactions from different representation spaces jointly.
%\begin{align}
%    &\mathrm{MultiHead}(H) =  [ \text{head}_1;...;\text{head}_{k} ] W^O,\\
%    &\quad \text{head}_i = \mathrm{Attn}(H W^Q_i,H W^K_i,H W^V_i),
%\end{align}
%where $W^O,W^Q_i,W^K_i,W^V_i (i\in [1,k])$ are learnable parameters.

The Transformer encoder consists of several stacked multi-head self-attention layers and fully-connected layers.  Assuming the input of the multi-head self-attention layer is $H$, its output $\tilde{H}$ is calculated by
\begin{align}
    Z =& \textrm{layer-norm}\Big(H + \mathrm{MultiHead}(H)\Big), \\
    \tilde{H} =& \textrm{layer-norm}\Big(Z + \mathrm{FFN}(Z)\Big), \label{eq:residue}
\end{align}
where $\textrm{layer-norm}(\cdot)$ represents the layer normalization \cite{ba2016layer} .

%Besides, since the self-attention ignores the order information of a sequence, a positional embedding $PE$ is used to represent the positional information.

All the tasks with the different criteria use the same encoder. Nevertheless, with the different criterion-token $[m]$, the encoder can effectively extract the criterion-aware representation for each character.

\subsection{Decoding Layer} In the standard multi-task learning framework, each task has its private decoder to predict the task-specific labels.
Different from the previous work, we use a shared decoder for all the tasks since we have extracted the criterion-aware representation for each character. In this work, we use CRF as the decoder since it is slightly better than MLP (see Sec. \ref{sec:ablation}).

With the fully-shared encoder and decoder, our model is more concise than the shared-private architectures \cite{chen2017adversarial,huang2019toward}.

% to minimize the cross-entropy when MLP is utilized or the negative log-likelihood when CRF is utilized.

\section{Experiments} \label{experiments}

\begin{table*}[t]\small \setlength{\tabcolsep}{7pt}
\centering
\caption{Details of the eight datasets after preprocessing. ``Word Types'' represents the number of unique word. ``Char Types'' is the number of unique characters. ``OOV Rate'' is Out-Of-Vobulary rate.}
 \label{tab:info_datasets}%$N_w$ and $N_c$ indicate numbers of tokens and characters respectively. $\mathcal{D}_w$ and $\mathcal{D}_c$ are the dictionaries of distinguished words and characters respectively. $N_s$ indicates the number of sentences.
       \begin{tabular}{ccrrrrrr}
    \toprule
           \multicolumn{3}{c}{Corpora} & Words\# & Chars\# & Word Types & Char Types & OOV \\
    \midrule
    \multirow{8}[8]{*}{\rotatebox{90}{Sighan05}} & \multirow{2}[2]{*}{MSRA} & Train & 2.4M  & 4.0M  & 75.4K & 5.1K  &  \\
          &       & Test  & 0.1M  & 0.2M  & 11.9K & 2.8K  & 1.32\% \\
\cmidrule{2-8}          & \multirow{2}[2]{*}{AS} & Train & 5.4M  & 8.3M  & 128.8K & 5.8K  &  \\
          &       & Test  & 0.1M  & 0.2M  & 18.0K & 3.4K  & 2.20\% \\
\cmidrule{2-8}          & \multirow{2}[2]{*}{PKU} & Train & 1.1M  & 1.8M  & 51.2K & 4.6K  &  \\
          &       & Test  & 0.1M  & 0.2M  & 12.5K & 2.9K  & 2.06\% \\
\cmidrule{2-8}          & \multirow{2}[2]{*}{CITYU} & Train & 1.1M  & 1.8M  & 43.4K & 4.2K  &  \\
          &       & Test  & 0.2M  & 0.4M  & 23.2K & 3.6K  & 3.69\% \\
    \midrule
    \multirow{8}[8]{*}{\rotatebox{90}{Sighan08}} & \multirow{2}[2]{*}{CTB} & Train & 0.6M  & 1.0M  & 40.5K & 4.2K  &  \\
          &       & Test  & 0.1M  & 0.1M  & 11.9K & 2.9K  & 3.80\% \\
\cmidrule{2-8}          & \multirow{2}[2]{*}{CKIP} & Train & 0.7M  & 1.1M  & 44.7K & 4.5K  &  \\
          &       & Test  & 0.1M  & 0.1M  & 14.2K & 3.1K  & 4.29\% \\
\cmidrule{2-8}          & \multirow{2}[2]{*}{NCC} & Train & 0.9M  & 1.4M  & 53.3K & 5.3K  &  \\
          &       & Test  & 0.2M  & 0.2M  & 20.9K & 3.9K  & 3.31\% \\
\cmidrule{2-8}          & \multirow{2}[2]{*}{SXU} & Train & 0.5M  & 0.8M  & 29.8K & 4.1K  &  \\
          &       & Test  & 0.1M  & 0.2M  & 11.6K & 2.8K  & 2.60\% \\
    \bottomrule
    \end{tabular}%
\end{table*}
%\rotatebox{90}{Sighan05}}

{\bf Datasets}
We use eight CWS datasets from SIGHAN2005 \cite{emerson2005second} and SIGHAN2008 \cite{moe2008fourth}. Among them, the AS, CITYU, and CKIP datasets are in traditional Chinese, while the MSRA, PKU, CTB, NCC, and SXU datasets are in simplified Chinese. Except where otherwise stated, we follow the setting of \cite{chen2017adversarial,gong2018switch}, and translate the AS, CITYU and CKIP datasets into simplified Chinese. We do not balance the datasets and randomly pick 10\% examples from the training set as the development set for all datasets. Similar to the previous work \cite{chen2017adversarial}, we preprocess all the datasets by replacing the
continuous Latin characters and digits with a
unique token, and converting all digits, punctuation and Latin
letters to half-width to deal with the full/half-width mismatch between training and test set.

We have checked the annotation schemes of different datasets, which are just partially shared and no two datasets have the same scheme. According to our statistic, the averaged overlap is about 20.5\% for 3-gram and 4.4\% for 5-gram.

Table \ref{tab:info_datasets} gives the details of the eight datasets after preprocessing.
For training and development sets, lines are split into shorter sentences or clauses by punctuations, in order to make a faster batch.
% W All datasets are preprocessed by

% We use the standard measures of F1 scores to evaluate Chinese word segmentation.
%The precision of Chinese word segmentation (denoted as $P$) is calculated by the number of correctly segmented words versus the total number of segmented words. The recall of Chinese word segmentation (denoted as $R$) is computed by the number of correctly segmented words versus the total number of golden words. Then we get $F1$ score by $F1=2*P*R/(P + R)$.

{\bf Pre-trained Embedding}
Based on on~\cite{chen2015long,shao2017character,zhang2018simple}, n-gram features are of great benefit to Chinese word segmentation and POS tagging tasks. Thus we use unigram and bigram embeddings for our models. We first pre-train unigram and bigram embeddings on Chinese Wikipedia corpus by the method proposed in~\cite{ling2015two}, which improves standard word2vec by incorporating token order information. %For a sentence with characters ``abcd...'', the unigram sequence is ``a b c ...''; the bigram sequence is ``ab bc cd ...''. In the training phase of CWS, all pre-trained embeddings are fixed at the first 50 epochs and then updated during our experiments.

\begin{table}[t]\setlength{\tabcolsep}{8pt}\small
    \centering
    \begin{tabular}{l|c}
      \toprule
      Embedding Size $d$ & 100 \\
      Hidden State Size $d_{model}$ & 256\\
      Transformer Encoder Layers & 6\\
      Attention Heads & 4\\
%      Gradients Clip                      & 5  \\
      Batch Size                          & 256 \\
      Dropout Ratio                 & 0.2 \\
%      MLP Depth                         & 1 \\
      Warmup Steps                  & 4000 \\
      %Max epochs                  & 100 \\
      \bottomrule
    \end{tabular}
    \caption{Hyper-Parameter Settings} \label{table:hyper}
\end{table}

{\bf Hyper-parameters}
We use Adam optimizer~\cite{kingma2014adam} with the same warmup strategy as \cite{vaswani2017attention}.
The development set is used for parameter tuning.
% We use the CRF as the default decoder since it is slightly better than MLP (see Sec. \ref{sec:ablation}).
All the models are trained for 100 epochs. Pre-trained embeddings are fixed for the first 80 epochs and then updated during the following epochs. After each training epoch, we test the model on the dev set, and models with the highest $F1$ in the dev set are used in the test set. Table \ref{table:hyper} shows the detailed hyperparameters.

%Since the numbers of layers and attention heads are important, we list the performances with different settings in Table \ref{tab:1}. Based on their performances, we use six layers of self-attention, each layer with four attention heads.
%
%\begin{table}[ht]\small
%  \centering
%    \begin{tabular}{cccc}
%    \toprule
%    \# of Layer & \# of head & avg F1 \\
%    \midrule
%    2     & 4    & 96.69 \\
%    4     & 4    & 96.81 \\
%    6     & 4    & \bf{96.87} \\
%    2     & 8    & 96.68 \\
%    4     & 8    & 96.8 \\
%    6        & 8       & 96.84 \\
%    \bottomrule
%    \end{tabular}%
%    \caption{Average F1 values with different number of Transformer Encoder layers and heads are used.}
%  \label{tab:1}%
%\end{table}%

% Table generated by Excel2LaTeX from sheet 'main results'
\begin{table*}[t!]\setlength{\tabcolsep}{4pt}
  \centering\small%\renewcommand{\arraystretch}{0.8}
      \begin{tabular}{lcccccccccc}
    \toprule
    Models &       & MSRA  & AS    & PKU   & CTB   & CKIP  & CITYU & NCC   & SXU   & Avg. \\
    \midrule
    \multicolumn{11}{l}{\textbf{Single-Criterion Models}} \\
    \midrule
    %\multirow{4}[2]{*}{BiLSTM} & P     & 95.7  & 93.64 & 93.67 & 95.19 & 92.44 & 94    & 91.86 & 95.11 & 93.95 \\
    %      & R     & 95.99 & 94.77 & 92.93 & 95.42 & 93.69 & 94.15 & 92.47 & 95.23 & 94.33 \\
    Stacked BiLSTM \cite{ma2018state} & F  & 97.4 & \textbf{96.2} & 96.1 & 96.7 & - & \textbf{97.2} & - & - &- \\
    BiLSTM \cite{chen2017adversarial}    & F & 95.84 & 94.2 & 93.3 & 95.3 & 93.06 & 94.07 & 92.17 & 95.17 & 94.14 \\
Switch-LSTMs \cite{gong2018switch}        & F & 96.46 & 94.51 & 95.74 & \textbf{97.09} & 92.88 & 93.71 & 92.12 & 95.57 & 94.76 \\
Transformer Encoder        & F & \textbf{98.07} & {96.06} & \textbf{96.39} & 96.41 & \textbf{95.66} & {96.32} & \textbf{95.57} & \textbf{97.08} & \textbf{96.45} \\
\rowcolor[gray]{0.7} Transformer Encoder          & OOV   & \textbf{73.75} & 73.05 & {72.82} & {82.82} & {79.05} & {83.72} & {71.81} & {77.95} & {76.87} \\
\midrule
\multicolumn{11}{l}{\textbf{Multi-Criteria Models}} \\
    \midrule
%    Models &       & MSRA  & AS    & PKU   & CTB   & CKIP  & CITYU & NCC   & SXU   & Avg. \\
%    \midrule

%BERT$^\star$  \cite{huang2019toward} &F & 97.9&  96.6 &  96.6 &  97.6  &-&97.6 &-& 97.3&-\\\midrule
%    \multirow{4}[2]{*}{BiLSTM} & P     & 95.95 & 94.17 & 94.86 & 96.02 & 93.82 & 95.39 & 92.46 & 96.07 & 94.84 \\
%          & R     & 96.14 & 95.11 & 93.78 & 96.33 & 94.7  & 95.7  & 93.19 & 96.01 & 95.12 \\
BiLSTM \cite{chen2017adversarial}        & F & 96.04 & 94.64 & 94.32 & 96.18 & 94.26 & 95.55 & 92.83 & 96.04 & 94.98 \\
%          & OOV   & 71.6  & 73.5  & 72.67 & 82.48 & 77.59 & 81.4  & 63.31 & 77.1  & 74.96 \\
    %\midrule
%    \multirow{4}[2]{*}{Switch-LSTMs} & P     & 97.69 & 94.42 & \textbf{96.24} & \textbf{97.09} & 94.53 & 95.85 & 94.07 & 96.88 & 95.85 \\
%          & R     & 97.87 & 96.03 & 96.05 & \textbf{97.43} & 95.45 & 96.59 & 94.17 & 97.62 & 96.4 \\
 Switch-LSTMs \cite{gong2018switch}       & F & 97.78 & 95.22 & 96.15 & \textbf{97.26} & 94.99 & 96.22 & 94.12 & 97.25 & 96.12 \\
 Unified BiLSTM \cite{he2019effective} &F & 97.2 &95.4 & 96.0  & 96.7 &-& 96.1 &-& 96.4 &-\\ %\midrule
%          & OOV   & 64.2  & 77.33 & 69.88 & 83.89 & 77.69 & 73.58 & 69.76 & 78.69 & 74.38 \\
    %\midrule
%    \multirow{4}[2]{*}{Transformer Encoder} & P     & \textbf{98.03} & \textbf{96.84} & 95.88 & 96.79 & \textbf{96.92} & \textbf{97.03} & \textbf{95.85} & \textbf{97.52} & \textbf{96.86} \\
%          & R     & \textbf{98.06} & \textbf{96.05} & \textbf{96.95} & 97.18 & \textbf{96.11} & \textbf{96.78} & \textbf{96.24} & \textbf{97.69} & \textbf{96.88} \\
Our Unified Model          & F & \textbf{98.05} & \textbf{96.44} & \textbf{96.41} & 96.99 & \textbf{96.51} & \textbf{96.91} & \textbf{96.04} & \textbf{97.61} & \textbf{96.87} \\
\rowcolor[gray]{0.7} Our Unified Model          & OOV   & \textbf{78.92} & {76.39} & {78.91} & {87} & {82.89} & {86.91} & {79.3} & {85.08} & {81.92} \\
    \bottomrule
    \end{tabular}%
    \caption{Overall results on eight CWS datasets. F and OOV indicate the $F1$ score and OOV recall, respectively. The upper block consists of single-criterion models.
    Since Stacked BiLSTM \cite{ma2018state} is a strong SOTA model, the other comparable CWS models are omitted for brevity.
    The lower block consists of multi-criteria models. %The models with $^\star$ are not directly compared since they trained on ten datasets (including six overlapped datasets with us).
    }
      \label{tab:res}
\end{table*}%
\begin{table*}[ht]%\setlength{\tabcolsep}{4pt}
  \centering\small%\renewcommand{\arraystretch}{0.8}
  \begin{threeparttable}
      \begin{tabular}{lccccccccc}
        \toprule
            Models & MSRA & AS & PKU & CTB & CKIP & CITYU & NCC & SXU & Avg. \\
          \midrule
            Unified Model & \textbf{98.05} & \textbf{96.44} & \textbf{96.41} & \textbf{96.99} & 96.51 & \textbf{96.91} & \textbf{96.04} & \textbf{97.61} & \textbf{96.87} \\
            w/o CRF & 98.02 & 96.42 & 96.41 & 96.9 & \textbf{96.59} & 96.87 & 95.96 & 97.5 & 96.83 \\
            w/o bigram & 97.41 & 96 & 96.25 & 96.71 & 96 & 96.31 & 94.62 & 96.84 & 96.27 \\
            w/o pre-trained emb. & 97.51 & 96.06 & 96.02 & 96.47 & 96.22 & 95.99 & 94.82 & 96.76 & 96.23 \\
          \bottomrule
    \end{tabular}%
    \end{threeparttable}
    \caption{Ablation experiments. %The first line presents the results of our unified model. The following lines are results of separately removing a certain part.
    } \label{tb:ablation}
    \label{tab:ablation}
\end{table*}

\subsection{Overall Results}

Table \ref{tab:res} shows the experiment results of the proposed model on test sets of eight CWS datasets.

We first compare our Transformer encoder with the previous models in the \textit{single-criterion scenario}. The comparison is presented in the upper block of Table \ref{tab:res}. Since Switch-LSTMs \cite{gong2018switch} is designed form MCCWS, it is  just slight better than BiLSTM in single-criterion scenario.
Compared to the LSTM-based encoders, the Transformer encoder brings a noticeable improvement compared to \cite{chen2017adversarial,gong2018switch},
and gives a comparable performance to \cite{ma2018state}.
In this work, we do not intend to prove the superiority of the Transformer encoder over LSTM-based encoders in the single-criterion scenario. Our purpose is to build a concise unified model based on Transformer encoder for MCCWS.

In the \textit{multi-criteria scenario}, we compare our unified model with the BiLSTM \cite{chen2017adversarial} and Switch-LSTMs \cite{gong2018switch}. The lower block of Table \ref{tab:res} displays the contrast. Firstly, although different criteria are trained together, our unified model achieves better performance besides CTB. Compared to the single-criterion scenario, 0.42 gain in average $F1$ score is obtained by the multi-criteria scenario.
Moreover, our unified model brings a significant improvement of 5.05 in OOV recall.
Secondly, compared to previous MCCWS models, our unified model also achieves better average $F1$ score. Especially, our unified model significantly outperforms the unified BiLSTM \cite{he2019effective}, which indicates the Transformer encoder is more effective in carrying the criterion information than BiLSTM. The reason is that the Transformer encoder can model the interaction of the criterion-token and each character directly, while BiLSTM needs to carry the criterion information step-by-step from the two ends to the middle of the input sentence.
The criterion information could be lost for the long sentences.

There are about 200 sentences are shared by more than one datasets with different segmentation schemes, but it is not much harder to correctly segment them. Their F1 score is 96.84.

\begin{table*}[th]\setlength{\tabcolsep}{5pt}
  \centering\small%\renewcommand{\arraystretch}{0.8}
  \begin{tabular}{*{10}{c}}
  \toprule
      Models & MSRA & AS & PKU & CTB & CKIP & CITYU & NCC & SXU & Avg. F1 \\
  \midrule
    8Simp & \textbf{98.05} & 96.44 & 96.41 & \textbf{96.99} & 96.51 & \textbf{96.91} & 96.04 & \textbf{97.61} & \textbf{96.87} \\
    8Trad & 97.98 & 96.39 & 96.49 & \textbf{96.99} & 96.49 & 96.86 & 95.98 & 97.48 & 96.83 \\
    5Simp, 3Trad & 98.03 & \textbf{96.52} & \textbf{96.6} & 96.94 & 96.38 & 96.8  & 96.02 & 97.55 & 96.86 \\
    8 Simp, 8 Trad & 98.04 & 96.41 & 96.43 & \textbf{96.99} & \textbf{96.54} & 96.85 & \textbf{96.08} & 97.52 & 96.86 \\
    \bottomrule
  \end{tabular}
  \caption{Joint training on both the simplified and traditional Chinese corpus.}
  \label{tab:mix}
\end{table*}

\begin{figure}[th]
    \centering
    \vspace{-1em}
    \includegraphics[width=0.35\textwidth]{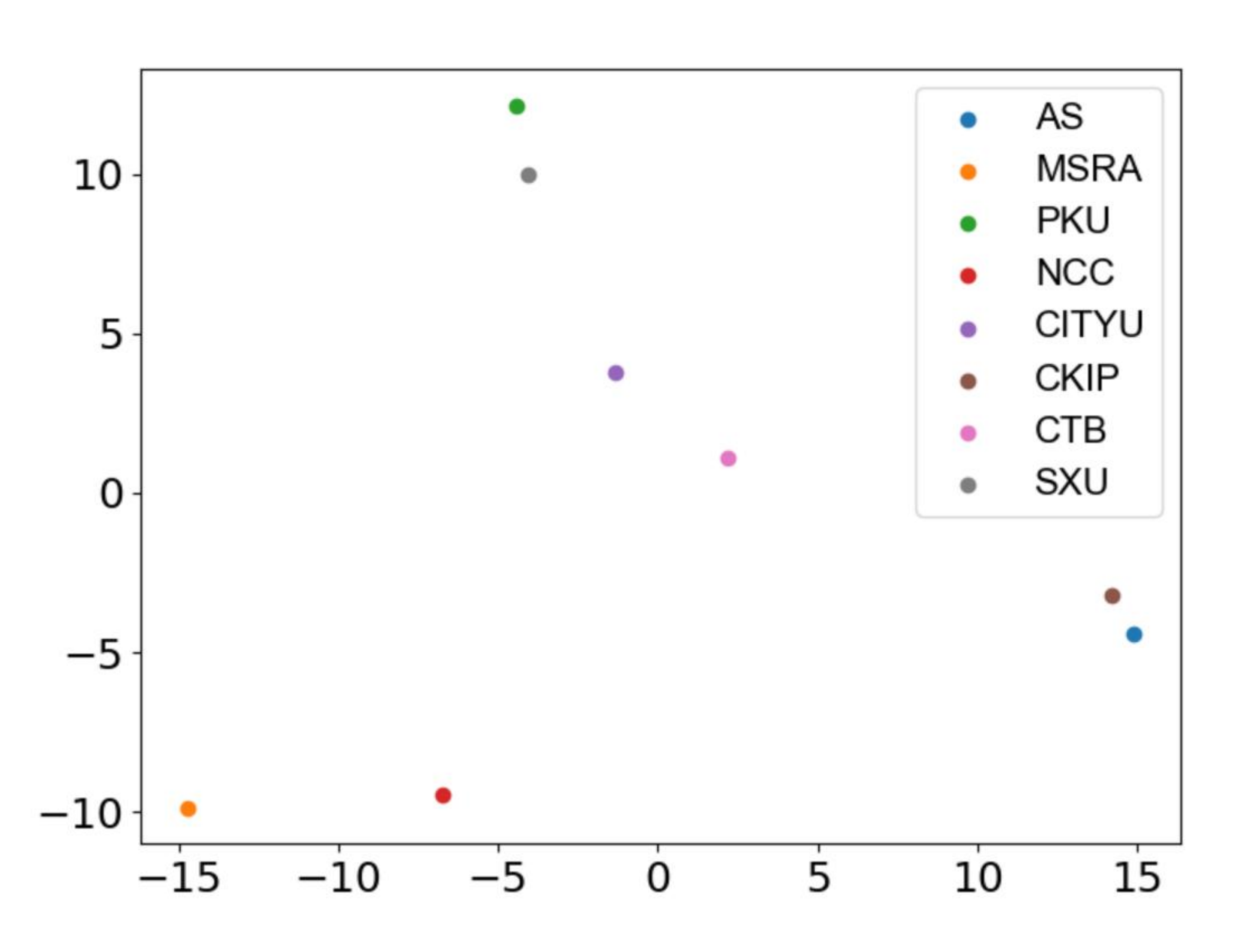}
    \vspace{-1em}
    \caption{Visualization of the criterion embeddings.}\label{fig:illus-emb}
  \end{figure}

Figure \ref{fig:illus-emb} visualizes the 2D PCA projection of the learned embeddings of eight different criteria. Generally, the eight criteria are mapped into dispersed points in the embedding space, which indicates that each criterion is different from others.
Among them, MSRA is obviously different from others. A possible reason is that the named entity is regarded as a whole word in the MSRA criterion, which is significantly distinguishing with other criteria.

\subsection{Ablation Study}\label{sec:ablation}

Table \ref{tab:ablation} shows the effectiveness of each component in our model.

% Table generated by Excel2LaTeX from sheet 'Sheet3'
\begin{table*}[h]\setlength{\tabcolsep}{3pt}\footnotesize
  \centering
    \begin{tabular}{llllll}
    \toprule
    苹果 (apple)   & \tradc{蘋果} (apple)    & 爱好(hobby)    & \tradc{愛好}(hobby)    & 担心(worry)    & \tradc{擔心}(worry) \\
    \midrule
    坚果(nut)    & \tradc{微軟}(Microsoft)    & 热爱(love)    & \tradc{熱愛}(love)    & 关心(care)    & \tradc{關心}(care) \\
    谷歌(Google) & \tradc{黃油 }(butter)  & 兴趣(interest)    & \underline{爱好}(hobby) & 怀疑(doubt)    & \tradc{顧慮}(misgiving)\\
    华为(Huawei)    & \tradc{現貨}(goods in stock)    & \underline{\tradc{愛好}}(hobby) & \tradc{興趣}(interest)    & 顾虑(misgiving)    & \tradc{懷疑}(doubt) \\
    黄油(butter)    & \tradc{果凍}(jelly)    & 梦想(dream)    & \tradc{夢想}(dream)    & 担忧(concern)    & \tradc{擔憂}(concern) \\
    鲜果(fresh fruit)    & \tradc{京東}(JD)    & 爱玩(Playful)    & \tradc{愛玩}(playful)    & 责怪(blame)    & \tradc{憂慮}(anxiety) \\
    微软(Microsoft)    & \tradc{賣家}(seller)    & 痴迷(addict)    & \tradc{喜愛}(adore)    & 伤心(sad)    & \tradc{責怪}(blame) \\
    诺基(Nokia)    & \underline{苹果}(apple) & 乐趣 (pleasure)   & \tradc{習慣}(habbit)    & 嫌弃(disfavour)    & \tradc{傷心}(sad) \\
    \underline{\tradc{蘋果}}(Apple) & \tradc{售後}(after-sales)    & 喜爱 (adore)   & \tradc{樂趣}(pleasure)    & 忧虑(anxiety)    & \underline{担心}(worry) \\
    \bottomrule
    \end{tabular}%
    \caption{Qualitative analysis for the joint embedding space of simplified and traditional
Chinese. Given the target bigram, we list its top 8 similar bigrams. The bigram with red color indicates it is traditional Chinese.}
  \label{tab:top10}%
\end{table*}%

The first ablation study is to verify the effectiveness of the CRF decoder, which is popular in most CWS models.
The comparison between the first two lines indicates that with or without CRF does not make much difference. Since a model with CRF takes a longer time to train and inference, we suggest not to use CRF in Transformer encoder models in practice.

The other two ablation studies are to evaluate the effect of the bigram feature and pre-trained embeddings. We can see that their effects vary in different datasets. Some datasets are more sensitive to the bigram feature, while others are more sensitive to pre-trained embeddings. In terms of average performance, the bigram feature and pre-trained embeddings are important and boost the performance considerably, but these two components do not have a clear winner.

\subsection{Joint Training on both simplified and Traditional Corpora}
\label{sec:bothlan}

In the above experiments, the traditional Chinese corpora (AS, CITYU, and CKIP) are translated into simplified Chinese.
However, it might be more attractive to jointly train a unified model directly on the mixed corpora of simplified and traditional Chinese without translation. As a reference, the single model has been used to translate between multiple languages in the field of machine translation \cite{johnson2017google}.

To thoroughly investigate the feasibility of this idea, we study four different settings to train our model on simplified and traditional Chinese corpora.

\begin{enumerate*}
  \item The first setting  (``8Simp'') is to translate all the corpora into simplified Chinese. For the pre-trained embeddings, we use the simplified Chinese Wikipedia dump to pre-train the unigram and bigram embeddings. This way is the same as the previous experiments.
  \item The second setting (``8Trad'') is to translate all the corpora into traditional Chinese. For the pre-trained embeddings, we first convert the Wikipedia dump into traditional Chinese characters, then we use this converted corpus to pre-train unigram and bigram embeddings.
  \item The third setting (``5Simp, 3Trad'') is to keep the original characters for five simplified Chinese corpora and three traditional Chinese corpora without translation. The unified model can take as input the simplified or traditional Chinese sentences. In this way, we pre-train the joint simplified and traditional Chinese embeddings in a joint embedding space. We merge the Wikipedia corpora used in ``8Trad'' and ``8Simp'' to form a mixed corpus, which contains both the simplified and traditional Chinese characters. The unigram and bigram embeddings are pre-trained on this mixed corpus.
  \item  The last setting (``8Simp, 8Trad'') is to simultaneously train our model on both the eight simplified Chinese corpora in ``8Simp'' and the eight traditional Chinese corpora in ``8Trad''. The pre-trained word embeddings are the same as ``5Simp, 3Trad''.
\end{enumerate*}

Table \ref{tab:mix} shows that there does not exist too much difference between different settings. This investigation indicates it is feasible to train a unified model directly on two kinds of Chinese characters.

To better understand the quality of the learned joint embedding space of the simplified and traditional Chinese, we conduct a qualitative analysis to illustrate the most similar bigrams for a target bigram. Similar bigrams are retrieved based on the cosine similarity calculated using the learned embeddings. As shown in Table \ref{tab:top10}, the traditional Chinese bigrams are similar to their simplified Chinese counterparts, and vice versa. The results show that the simplified and traditional Chinese bigrams are aligned well in the joint embedding space.

\subsection{Transfer Capability}
\label{sec:transfer}

Since except for the criterion embedding, the other parts of the unified model are the same for different criteria, we want to exploit whether a trained unified model can be transferred to a new criterion only by learning a new criterion embedding with few examples.

\begin{figure}[t!]
    \centering
    \begin{tikzpicture}[scale=0.8]
      \begin{axis}[
  %    title=Inv. cum. normal,
    %width = 0.5\linewidth,height = 0.5\linewidth,
      %font=\scriptsize,
      xlabel={Number of Training Samples},
      %xtick=data,
      %xticklabels={1,2,3,4,5,6,7,8,9,10,32,64},
      %ymin=88,
      %xtick distance=0.2,
      %yticklabel style={
       %   font=\tiny
      %},
      ylabel={Averaged $F1$ score},
      %ylabel style={
      %    yshift=-1ex,
      %},
      %mark size=1.0pt,
      %no markers,
    %   ymajorgrids=true,
    %   grid style=dashed,
      %cycle list name=exotic,
     %cycle list name=my black white,
      legend cell align={left},
      legend entries={Ours$^{(trans)}$,Switch-LSTMs$^{(trans)}$,Ours,Switch-LSTMs},
      %legend to name=named,
      legend pos= south east,
      %legend style={font=\tiny,line width=.5pt,mark size=.5pt,
%              %at={(2,1.1)},
%              anchor=east,
%              legend columns=-1,
%              %legend rows=2,
%              /tikz/every even column/.append style={column sep=0.5em}},
%              smooth,
      ]
      \addplot [red,mark=*] table [x=num, y=transt] {./transfer.txt};
      \addplot [green,mark=*] table [x=num, y=lstmt] {./transfer.txt};
      \addplot [blue!50!red,mark=*] table [x=num, y=trans] {./transfer.txt};
      \addplot [blue,mark=*] table [x=num, y=lstm] {./transfer.txt};
      \end{axis}
  \end{tikzpicture}
  \caption{Evaluation of the transfer capability. Switch-LSTMs and Ours are models trained on the given instances from scratch. Switch-LSTMs$^{(trans)}$ and Ours$^{(trans)}$ are models learned in transfer fashion.}
  \label{fig:transfer}
  \end{figure}
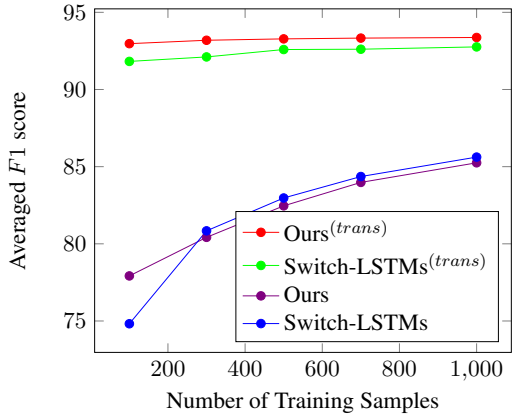

We use the leave-one-out strategy to evaluate the transfer capability of our unified model. We first train a model on seven datasets, then only learn the new criterion embedding with a few training instances from the left dataset. This scenario is also discussed in \cite{gong2018switch}, and Figure \ref{fig:transfer} presents their and our outcomes (averaged $F1$ score).
There are two observations: Firstly, for the different number of samples, the transferred model always largely outperforms the models learned from scratch. We believe this indicates that learning a new criterion embedding is an effective way to transfer a trained unified model to a new criterion. Secondly, our model also has superior transferability than Switch-LSTMs (Ours$^{(trans)}$ \textit{versus} Switch-LSTMs$^{(trans)}$).
%, since the average $F1$ scores with the different number of target samples are better than its Switch-LSTMs counterparts.

\section{Related Work}

%Multi-task learning utilizes the correlation between related tasks to improve classification by learning tasks in parallel.

The previous work on the MCCWS can be categorized into two lines.

One line is multi-task based MCCWS. \citet{chen2017adversarial} proposed a multi-criteria learning framework for CWS, which uses a shared layer to extract the common underlying features and a private layer for each criterion to extract criteria-specific features.
\citet{huang2019toward} proposed a domain adaptive segmenter to capture diverse criteria based on  Bidirectional Encoder Representations from Transformer (BERT) \cite{DBLP:journals/corr/abs-1810-04805}.

Another line is unified MCCWS. \citet{gong2018switch} presented Switch-LSTMs to segment sentences, which consists of several LSTM layers, and uses a criterion switcher at every position to change the routing among these LSTMs automatically. However, the complexity of the model makes Switch-LSTMs hard to be applied in practice. \citet{he2019effective} used a shared BiLSTM by adding two artificial tokens at the beginning and end of an input sentence to specify the output criterion. However, due to the long-range dependency problem, BiLSTM is hard to carry the criterion information to each character in a long sentence.

Compared to the above two unified models,  we use the Transformer encoder in our unified model, which can elegantly model the criterion-aware context representation for each character.
With the Transformer, we just need a special criterion-token to specify the output criterion. Each character can directly attend the criterion-token to be aware of the target criterion.
Thus, we can use a single model to produce different segmented results for different criteria. Different from \cite{huang2019toward}, which uses the pre-trained Transformer BERT and several extra projection layers for different criteria, our model is a fully-shared and more concise.

\section{Conclusion and Future Work}
We propose a concise unified model for MCCWS, which uses the Transformer encoder to extract the criterion-aware representation according to a unique criterion-token.  Experiments on eight corpora show that our proposed model outperforms the previous models and has a stronger transfer capability. The conciseness of our model makes it easy to be applied in practice.

In this work, we only adopt the vanilla Transformer encoder since we just want to utilize its self-attention mechanism to model the criterion-aware context representation for each character neatly. Therefore, it is promising for future work to look for the more effective adapted Transformer encoder for CWS task or to utilize the pre-trained models \cite{qiu2020:scts-ptms}, such as BERT-based MCCWS \cite{ke2020unified}.
Besides, we are also planning to incorporate other sequence labeling tasks into the unified model,
such as POS tagging and named entity recognition.

\section{Acknowledgements}
This work was supported by the National Natural Science Foundation of China (No. 62022027 and 61976056), Science and Technology on Parallel and Distributed Processing Laboratory (PDL).
\end{CJK*}

\bibliographystyle{acl_natbib}
\bibliography{nlp}

\end{document}